\documentclass[11pt]{article}
\usepackage[margin=1in]{geometry}
\usepackage{hyperref}
\usepackage{booktabs}
\hypersetup{hidelinks}
\title{GroundBench: A Factorized, Counterfactual Benchmark\\
for Locating VLM Affordance Failures}
\author{Sarthak Sattigeri\\
\small Manipal University Jaipur\\
\small \url{https://github.com/S1rlvk/groundbench}}
\date{}

\begin{document}
\maketitle

\begin{abstract}
A companion evaluation found that naming the target part in a manipulation prompt raises action accuracy by 0.32 to 0.63 across eight vision-language models, with no model beating a constant baseline until the part is named. Naming the part, however, supplies a variable a real system must still produce; the comparison therefore cannot distinguish visual grounding, mechanical reasoning, and category-to-action association. We introduce GroundBench to separate these explanations. Its six conditions form a branch-and-merge design in which each edge adds one controlled information bundle, and a counterfactual re-ask names a real noncanonical part visible in the same image. We evaluate three OpenAI models on 1{,}068 predictions. Supplying the benchmark's target region while withholding its identity leaves all three models at or below the 0.53 majority baseline (0.26, 0.26 and 0.53), despite near-exact reproduction of the supplied region. Supplying target identity without location instead yields 0.74, 0.68 and 0.68. Every above-baseline gain occurs where the supplied target-part category alone determines the action in this curated set. A no-vision control costs GPT-5 nothing on these conditions and improves two of three scores, evidence consistent with category-to-action association contributing substantially to the companion result; GPT-4o mini drops on one condition, leaving that interpretation non-universal. Adding joint type and motion axis never helps across six model-stratum comparisons. GPT-5 follows the noncanonical request on 0.86 of counterfactual pairs at a 0.07 shortcut rate, but fails every \texttt{push} to \texttt{lift-vertical} case. GroundBench thus identifies which supplied information changes affordance behavior while exposing when apparently grounded performance can be reproduced from text.
\end{abstract}

\vspace{0.4em}

\section{Introduction}

Asking a vision-language model what a robot should do with an object bundles at least three separable questions: which part matters, where that part is, and what happens if you act on it. A companion evaluation isolated the first two of these by comparing an open prompt, in which the model chooses which part to discuss, against a prompt that names the target part outright. The gap between them was large and consistent: across eight models from three developers, naming the part raised action accuracy by 0.32 to 0.63, and recovered the \texttt{push} action almost completely, from a single correct prediction across 64 model-object evaluations to 7 or 8 out of 8 for seven of the eight models. Under the open prompt, no model beat a constant answer that ignores the image; once the part was named, all eight did.

That comparison is informative and also incomplete by construction. Naming the part in the prompt is not a measurement of a part detector, it is a substitute for one. The result tells us what a pipeline could achieve if part identification were solved perfectly upstream of the language model. It does not tell us whether any of the eight models can solve that upstream problem themselves, whether the improvement reflects real mechanical reasoning about the named part or a simpler match between an object category and its typical action, or what happens to accuracy as the amount of supplied information is varied more finely than a single before-and-after step.

GroundBench is built to ask those remaining questions directly rather than infer them from a two-point comparison. It replaces the earlier evaluation's open-prompt-versus-named-part design with six conditions: nothing beyond the image, the object's category, the target part's identity, the part's image region with its identity withheld, that region together with its identity, and a bundled mechanics descriptor containing joint type and motion axis. Identity and region are added independently on top of the object category, not one after the other, so a model given the region without the identity is not partway down a single sequence toward the identity; it is on the other branch. Each edge changes one controlled information bundle while holding the other prompt fields fixed. A separate counterfactual construction supplies a real but noncanonical part name from the same image and object. It tests whether the model's output follows the changed part request or defaults to the original target's action; because the part name itself may imply an action, this is a prompt-compliance test rather than proof of visual attention.

This paper describes that construction: the six-condition factorized design, the data sources and how they are audited, the counterfactual and free-response variants, and the metrics and baselines built to score them. We built explicit safeguards against two measurement errors the companion evaluation only caught by testing its own numbers against a constant baseline after the fact, a threshold that let a baseline ignoring the image score 0.929, and an action label rule wrong on 4 of 19 objects. Every threshold and baseline in GroundBench's evaluator is derived from the data being scored rather than fixed in advance, specifically so a result cannot look strong only because the check for a trivial explanation was never run. Section~\ref{sec:results} reports what the design finds when run against three OpenAI models: the condition that supplies the benchmark's target region without a part name is worth nothing above a constant baseline, while gains occur only in conditions where the supplied target-part category already determines the answer within the curated set.

\section{Related work}

GroundBench follows a diagnostic rather than leaderboard-only view of evaluation. CLEVR separates visual-reasoning abilities that aggregate VQA scores conflate \cite{clevr}; CheckList organizes behavioral tests around capabilities and controlled perturbations \cite{checklist}; and sanity checks for saliency methods demonstrate that a plausible-looking metric can remain insensitive to the model or data it is meant to explain \cite{sanity}. These are instances of the broader shortcut-learning problem: a system can perform well by exploiting an unintended but predictive rule \cite{shortcut}. GroundBench applies the same methodological principle to part-level affordance prompts by varying supplied information and reporting trivial baselines alongside model scores.

Language priors are a particularly relevant shortcut in multimodal evaluation. VQA v2 pairs similar images with different answers to the same question so that language alone is insufficient \cite{vqav2}; VQA-CP changes answer priors between train and test \cite{vqacp}; and RUBi explicitly measures and suppresses question-only bias \cite{rubi}. Winoground similarly uses controlled image-caption pairs with identical words in different arrangements to probe whether success reflects cross-modal composition \cite{winoground}. GroundBench's no-vision control and within-object action-contrast re-ask inherit this logic, but target the distinction between a supplied part category, its image region, and the action label.

Referring-expression benchmarks such as ReferItGame evaluate language used to identify regions in natural images \cite{referit}, while MDETR learns text-conditioned detection and is evaluated on phrase grounding and referring-expression comprehension \cite{mdetr}. GroundBench does not propose a new grounding architecture. Instead, it asks a downstream diagnostic question those localization benchmarks do not: once a region or its semantic identity is supplied, which one changes affordance selection, and can the resulting score be obtained without the image?

Embodied language and vision-language systems connect semantic knowledge to action at several levels. SayCan constrains language-model plans with learned robotic affordances \cite{saycan}; PaLM-E incorporates continuous sensor inputs into an embodied multimodal language model \cite{palme}; and RT-2 maps vision-language inputs directly to tokenized robot actions \cite{rt2}. ManipBench evaluates lower-level manipulation reasoning and reports articulated-object manipulation as its hardest category, at 0.396 mean accuracy across 33 model variants against roughly 0.99 for humans \cite{manipbench}. Work on non-humanoid robot morphologies also reports substitution of common actions for correct affordances \cite{nonhumanoid}. GroundBench narrows the unit of analysis further, separating target identity, target region, and action choice within a single articulated-object prompt.

Prior affordance work provides complementary supervision and tasks. The UMD Part Affordance dataset supplies real tool images with part-level affordance annotations \cite{umd}; Where2Act predicts localized interaction regions and trajectories for articulated objects \cite{where2act}; GAPartNet supplies generalizable actionable-part categories \cite{gapartnet}; and AffordanceLLM uses VLM knowledge for affordance grounding in previously unseen categories \cite{affordancellm}. SAPIEN and PartNet-Mobility provide the articulated simulation substrate used by several of these lines of work \cite{partnetmobility}. None of these datasets was designed around a factorized prompt intervention or a within-image action-contrast re-ask. GroundBench contributes those controls rather than a replacement affordance dataset or policy.

\section{Benchmark design}
\label{sec:design}

\subsection{A six-condition factorized design}

Each example pairs an image with one or more annotated candidate parts, one of which is the target, and asks a model to name the actionable part, the action it affords, and where it is. GroundBench evaluates every example under a subset of six conditions:

\begin{itemize}
\item[] \texttt{image\_only}: no information beyond the image itself.
\item[] \texttt{object\_named}: the object's category is given.
\item[] \texttt{part\_named}: the object category and target part id/category are given.
\item[] \texttt{localization\_only}: the object category and target region are given, as a point and bounding box, while the target part id/category are withheld.
\item[] \texttt{localization\_shown}: the object category, target part id/category, and target region are given.
\item[] \texttt{mechanics\_shown}: a bundled mechanics descriptor (joint type and motion axis) is added to \texttt{localization\_shown}.
\end{itemize}

These six conditions are not one single chain: from \texttt{object\_named}, they branch. \texttt{part\_named} and \texttt{localization\_only} independently add one different information bundle to \texttt{object\_named}, target-part identity in one case and target region in the other. \texttt{localization\_shown} is where the branches merge, and \texttt{mechanics\_shown} adds the joint-type-and-axis descriptor. Thus the structure is \texttt{image\_only} $\to$ \texttt{object\_named} $\to$ \{\texttt{part\_named}, \texttt{localization\_only}\} $\to$ \texttt{localization\_shown} $\to$ \texttt{mechanics\_shown}: nested along each edge, not end to end as a single sequence. We treat part id/category and joint type/axis as predefined bundles; the design does not identify the effects of fields within either bundle. \texttt{object\_named} and \texttt{part\_named} correspond respectively to the open and named-part prompts in the companion evaluation; the remaining three are new and were built to measure what that comparison could not.

For brevity, ``target identity'' below denotes the supplied part id/category bundle, and ``target region'' denotes the supplied point/bounding-box bundle. Neither term implies that its component fields have been separately identified.

\texttt{localization\_only} is the control this branching makes possible, and it is not simply a weaker \texttt{localization\_shown}. \texttt{localization\_shown} hands over a region \emph{and} the target part's id/category, so any gain relative to \texttt{object\_named} could come from either bundle. \texttt{localization\_only} withholds the target part's id/category while keeping the region, and \texttt{part\_named} does the reverse, so the pair separates knowing \emph{where} the target part is from knowing \emph{what it is called}. Both retain the object-category line already supplied by \texttt{object\_named}; consequently, each differs from that common predecessor by one bundle. This distinction is the one the companion evaluation's named-part prompt could not draw at all, and Section~\ref{sec:results} shows the two bundles have very different measured associations with accuracy.

Neither \texttt{image\_only} nor \texttt{object\_named} supplies a goal, and an object can have more than one genuinely actionable part: a washing machine has both a door and buttons, each a real, correct answer to a different intended task. Scoring these two conditions against one hidden canonical part would conflate a model failing to find any part with one choosing a different, equally valid part for a task the prompt never specified, the exact ambiguity the companion evaluation's own open prompt carried. We score them instead as task-free discovery: a prediction counts as a valid part selection if it names any real annotated part whose action is defined, and the predicted action is checked against that part's own action rather than the hidden target's. A model that names the door and correctly says \texttt{pull} is scored correct here, even though this benchmark's own audit designated a button as the canonical target for that object. In conditions where the target is established explicitly by its identity or supplied region, this task-free ambiguity no longer applies and scoring uses the designated target. We separately report, for every condition, whether a prediction lands on the same specific part the audit designated, which coincides with the ordinary part-selection figure once the target identity is supplied and is a stricter, usually lower number before that. An alternative fix, adding an explicit goal that stays fixed across the whole design (for instance, ``turn the camera on'') and inserting a seventh condition between \texttt{object\_named} and \texttt{part\_named}, would match the causal story more precisely but requires writing a real goal for every object, phrased abstractly enough not to name the part or action outright; we defer that as a larger extension rather than build it here.

\subsection{Data sources and strata}

GroundBench draws on two source datasets, each split into an audited and an unaudited stratum that are never pooled into a single count without saying so. The GAPartNet stratum (1{,}046 raw articulated objects, automatically converted where a usable render and part annotation exist) yields 4{,}133 records across 600 objects; its action distribution is dominated by \texttt{push} (2{,}978 records, nearly all individually annotated keyboard keys and remote buttons) over \texttt{pull} (1{,}096) and \texttt{lift-vertical} (59), and every record in this stratum uses a target-highlighted render rather than a clean one, which makes it usable for \texttt{localization\_shown} and above but not for \texttt{image\_only} or \texttt{object\_named} without leaking the answer. A curated stratum of 19 GAPartNet objects, the same 19 individually spot-checked in the companion evaluation, uses clean, non-highlighted renders instead; 17 of the 19 have at least one other annotated part on the same object available as a real counterfactual distractor, and the remaining two, a printer's single button and a dishwasher's single door, do not.

The UMD stratum contributes 58{,}478 records at one per video frame across roughly 105 real tool instances, and a curated stratum of the same 28 objects used in the companion evaluation's spatial task, each reduced to one representative frame with a grasp point verified against the dataset's own segmentation mask rather than approximated. All 28 curated UMD objects carry at least one real non-target affordance region on the same object (26 have exactly one, such as a hammer's head alongside its handle; two mugs have two), which the bulk, per-frame stratum does not guarantee as reliably across every frame.

\subsection{Counterfactual construction}

A part-selection accuracy computed only over correctly identified targets cannot distinguish a model that recognizes the part it was asked about from one that recognizes the object and answers with that object's typical action regardless of which part was named. GroundBench's counterfactual generator addresses this directly rather than through a different accuracy computation: for every example with two or more annotated candidate parts, it emits a derived example per non-target part, keeping the same image and candidate parts but declaring one noncanonical part as the requested target. The original benchmark target remains recorded in the derived example's metadata, so a model that answers with the original target's action instead of the noncanonical part's action produces a distinguishable, reportable shortcut outcome rather than a hidden success. We call this a counterfactual re-ask throughout; it changes the requested target, not the image. Applied to the 19 curated GAPartNet objects, this construction yields 247 counterfactual examples from the 17 qualifying objects, ranging from a single swap on objects with only two annotated parts to 59 on a keyboard whose many individually annotated keys each count as a separate real distractor.

This is a prompt-level construction: the image, the object, and the set of real candidate parts are unchanged, only which part is declared the target changes. It does not composite a part from a different object into the scene, and it does not edit mesh geometry to substitute a differently-articulated part into the same location; both are real, more expensive extensions we have deliberately deferred rather than attempted here.

\subsection{Action-contrast mining}
\label{sec:action-contrast}

The exhaustive re-ask above only tests action compliance when the noncanonical part's action differs from the original benchmark target's. Scoring its 247 examples against that criterion found just 19 (7.7\%) informative, because most candidate parts on one object share a single action, a keyboard's dozens of keys all being \texttt{push} being the extreme case. Rather than re-ask exhaustively and hope some pairs turn out informative, we built a second construction that mines directly for objects containing at least two distinct, defined action labels among their annotated parts.

Applied to the bulk GAPartNet stratum (600 objects), only 8 qualify. Applied to the remaining roughly 446 raw objects that never passed the bulk converter's render-availability filter, 49 more qualify, every one of them lacking any render at all. We rendered 47 of these with the same offscreen renderer used for the curated stratum (2 already had renders from that earlier build), excluded 8 for the same closed-pose ambiguity that build's own methodology caught, and visually verified the projected bounding box of every remaining candidate part against its actual render, 150 candidate-part projections in total. 24 failed, the box landing off the object, on a featureless area, or on the object's own silhouette rather than a real part boundary, and were excluded before any model evaluation.

The original 8 bulk-stratum objects turned out to be unusable for a second, independent reason: their images are inherited from that stratum's per-target highlighted renders, which show the \emph{original} target highlighted rather than the \emph{swapped} one the record declares, so the visual cue and the declared target disagree. Rendering these 8 clean did not rescue them either; applying the same verification bar to their contrast parts left only one object's pair intact. All 8 objects and their 14 derived pairs are retained in the released data as a documented record of both findings, explicitly tagged as excluded from evaluation, not used in any figure below.

The frozen action-contrast pool is 32 objects and 74 pairs, entirely clean renders with verified exact boxes. Once model predictions exist we report three figures rather than one: pair-weighted compliance (every pair counted once), object-macro compliance (every object's own mean counted once, so an object contributing several pairs cannot outweigh one contributing a single pair), and a 95\% bootstrap interval resampled by object rather than by pair, since pairs from the same object are correlated evidence, not independent trials. A transition table records compliance and shortcut rates separately for each observed (original action, counterfactual action) pair, so a low pooled figure driven by one especially hard transition direction is visible rather than hidden.

\subsection{Scoring a free-form response}

The companion evaluation also asked whether a fixed label set was itself suppressing an answer a model would otherwise give, by comparing accuracy under a constrained label set against free prose describing the same named part, scored against keyword sets fixed before seeing any output. GroundBench reproduces this as a variant of \texttt{part\_named} rather than as a sixth condition, since it changes the answer's format rather than the information supplied to produce it: a prediction may carry free-form text scored against an action's keyword set instead of a label drawn from the fixed action vocabulary. The keyword table records, in the same file that defines it, that its \texttt{push} entry is deliberately more permissive than the others, matching a bias the companion evaluation disclosed in its own scoring rather than treating that asymmetry as invisible.

\subsection{Metrics and measurement safeguards}

For every scored condition, GroundBench reports part-selection accuracy, action accuracy, two conditional action accuracies, and two localization measures, evaluated against two baselines that ignore the image entirely.

The two conditional accuracies answer different questions and are reported separately rather than conflated: one restricts to predictions that named the exact designated part, the other to predictions whose point fell within a data-derived distance of the audited target regardless of what part name was given. A model can satisfy one without the other, naming the right part while pointing at the wrong spot, or pointing correctly while misnaming the part, and a single conditional accuracy would hide whichever failure mode it wasn't built to catch.

The companion evaluation's own reported errors motivate the localization safeguard specifically. A hand-picked distance threshold of 0.15 of the image diagonal, chosen before any model output existed, turned out to be looser than how tightly the underlying datasets center their ground truth, so a constant answer that never looks at the image scored a 0.929 hit rate under that threshold and made three real models look nearly as good as a baseline that ignores the image. GroundBench derives its localization threshold from the spread of the target centers actually being evaluated, and reports what the same constant baseline scores at that same threshold alongside every real hit-rate result, so a repeat of that error would be visible in the same row rather than requiring a separate check months later. An object-center baseline, distance from an object's own visual center rather than the image's center, is reported alongside the simpler image-center baseline for the same reason: an object rarely fills the frame symmetrically, and a model that does no better than pointing at wherever the object roughly is should not be mistaken for one reasoning about which part within it matters.

\section{Results}
\label{sec:results}

We evaluated three OpenAI models, GPT-4o mini, GPT-5 mini and GPT-5, across the two curated strata and the frozen action-contrast pool: 1{,}068 scored predictions for \$4.17, with every requested prediction returned and no unusable responses. Section~\ref{sec:no-vision} reports a no-vision control on the same models; Section~\ref{sec:qwen-pilot} reports a separate open-weight pilot whose main value is a measurement lesson rather than a result.

\subsection{A location is not a name}
\label{sec:location-not-name}

\begin{table}[h]
\centering
\small
\begin{tabular}{lccc c}
\toprule
Condition & GPT-4o mini & GPT-5 mini & GPT-5 & Answer in supplied text? \\
\midrule
\texttt{image\_only}         & 0.05 & 0.16 & 0.26 & no \\
\texttt{object\_named}       & 0.26 & 0.26 & 0.42 & no \\
\texttt{localization\_only}  & 0.26 & 0.26 & 0.53 & no \\
\midrule
\texttt{part\_named}         & 0.74 & 0.68 & 0.68 & yes \\
\texttt{localization\_shown} & 0.84 & 0.68 & 0.89 & yes \\
\texttt{mechanics\_shown}    & 0.68 & 0.53 & 0.79 & yes \\
\midrule
Majority-action baseline     & 0.53 & 0.53 & 0.53 & --- \\
\bottomrule
\end{tabular}
\caption{Action accuracy on the 19 audited GAPartNet objects, scored uniformly against the audited target's action in every condition. The evaluator also reports a selection-gated variant, which credits an action only once the model has named a valid part, and we use the ungated measure here for two reasons. The row every condition is compared against is a constant answer that ignores the image and selects no part at all, so gating the models on a selection the baseline never has to make is not a like-for-like comparison. Moreover, the gate applies only to the two task-free conditions: in the other four, the prompt establishes the target through its identity, its supplied region, or both. A gated column would therefore measure one quantity for the task-free rows and another for the target-established rows. The rightmost column is the evaluator's own leakage check, computed over the examples being scored rather than asserted: it reports whether every part sharing a category also shares an action, so that the category string handed to the model in that condition determines the answer without reference to the image. No condition in the upper block clears the constant-answer baseline for any model. \texttt{part\_named} and \texttt{localization\_shown} clear it for every model; \texttt{mechanics\_shown} clears it for two models and ties it for GPT-5 mini. Under the gated variant \texttt{image\_only} falls to 0.05, 0.05 and 0.00 and \texttt{object\_named} to 0.21, 0.05 and 0.16, no other row changes, and the comparison this section rests on lies entirely among rows it does not affect. The two readings are not ordered one above the other: on the curated UMD stratum the gated figure for GPT-5 under \texttt{image\_only} is 0.29 against 0.04 for the measure used here.}
\end{table}

The upper and lower blocks of the table divide exactly where the leakage check does, and the boundary is not the boundary the design was built around.

Consider \texttt{localization\_only} against \texttt{part\_named}. The first supplies the benchmark's target-region bounding box and point while withholding what the part is called; the second supplies the identity and no target region. On the same 19 objects, the same models, and the same action vocabulary, the region condition scores 0.26, 0.26 and 0.53 against a majority-action baseline of 0.53, and the identity condition scores 0.74, 0.68 and 0.68. Not one model exceeds the baseline when given the region alone, and GPT-5 lands on it exactly, 10 of 19 in both cases. Seven target boxes are verified exact projections and 12 are verified approximations, as detailed in Section~\ref{sec:limitations}, so this result concerns the benchmark-supplied region rather than a claim of perfect geometric ground truth.

This is not a failure to use the supplied region. Under \texttt{localization\_only} the models reproduce it almost exactly, with localization hit rates of 1.00, 0.84 and 0.95 and mean centre distances of 0.000, 0.025 and 0.012, against 0.21 for a constant answer at the same data-derived threshold. They take the box, put their point inside it, and still do not select the correct action more often than the constant baseline. In this sample, using the supplied region and mapping the target-part category to an action are separable behaviors, and only the second is associated with the companion evaluation's gain.

The leakage column offers a direct alternative explanation. Every above-baseline score in the lower block occurs under a condition containing a target-part category, such as \texttt{hinge\_door} or \texttt{slider\_button}, from which the action follows by lookup across the whole curated set. Thus a model could obtain these gains by reading the category and consulting a five-entry mapping rather than inspecting the image. Because the companion evaluation's headline gap of 0.32 to 0.63 crosses this same boundary, the present result is evidence consistent with text association contributing substantially to that gap; it does not estimate a unique causal share from this 19-object sample.

\subsection{The no-vision control}
\label{sec:no-vision}

We built a \texttt{text\_only} control that reissues every prompt with the image content block omitted entirely and ran it against all three models on the same 19 objects, 342 predictions for \$1.06. We scope this control to GAPartNet; the curated UMD stratum's majority-action baseline is already 1.00, so removing the image there cannot separate a model using it from one that cannot miss. We did not run a corresponding counterfactual no-vision control. Such a control would not preserve visual grounding, but it would reveal how much counterfactual compliance is achievable from the supplied part-category text alone; its absence limits the interpretation in Section~\ref{sec:counterfactual-results}.

\begin{table}[h]
\centering
\small
\begin{tabular}{lccc}
\toprule
Condition (leak flag) & GPT-4o mini & GPT-5 mini & GPT-5 \\
\midrule
\texttt{localization\_only} (no)      & 5 $\to$ 5   & 5 $\to$ 9   & 10 $\to$ 11 \\
\texttt{part\_named} (yes)            & 14 $\to$ 11 & 13 $\to$ 12 & 13 $\to$ 17 \\
\texttt{localization\_shown} (yes)    & 16 $\to$ 9  & 13 $\to$ 12 & 17 $\to$ 17 \\
\texttt{mechanics\_shown} (yes)       & 13 $\to$ 11 & 10 $\to$ 10 & 15 $\to$ 17 \\
\bottomrule
\end{tabular}
\caption{Correct out of 19, image-shown $\to$ text-only, on the same objects and prompts. Baseline (majority action) is 10 of 19.}
\end{table}

For GPT-5 the result is consistent with the leakage diagnosis. On every condition the check flags, removing the image costs nothing in this run, and the score rises from 13 to 17 on \texttt{part\_named} and from 15 to 17 on \texttt{mechanics\_shown}. \texttt{localization\_only}, the condition the previous subsection's headline claim rests on, moves from 10 to 11 of 19 with no image at all, with bootstrap intervals of $[0.32, 0.74]$ and $[0.37, 0.79]$. These intervals are descriptive rather than an equivalence test, but the direction is consistent with a model relying primarily on the supplied text in the leak-flagged conditions.

GPT-5 mini's shifts are small in both directions and its intervals overlap; we read this control as uninformative for that model rather than as evidence either way.

GPT-4o mini complicates the story rather than confirming it. \texttt{localization\_shown} drops from 16 of 19 to 9 of 19 with the image removed, with bootstrap intervals of $[0.68, 1.0]$ and $[0.26, 0.68]$. This is evidence that the image condition changes this model's behavior even though the target-part category is sufficient within the dataset. The model also wrapped nearly every otherwise usable answer in fenced JSON rather than returning the requested bare object: 113 of 114 image-shown responses violated the strict format, as did 66 of 114 text-only responses. Of the latter, one used an action outside the benchmark vocabulary and was normalized to \texttt{none}, making it incorrect; the other 65 were semantically recoverable fenced objects. This protocol noncompliance introduces a model-specific parsing path. With one sample per prompt and API-default decoding, this run cannot distinguish visual dependence, sampling variation, and any interaction with that recovery path.

The overall result is therefore mixed. GPT-5's result aligns with the leakage-check argument on every condition it was built to test; GPT-4o mini disagrees on one of the three. Replicated sampling or a paired test on a larger audited set is needed before treating either change as a stable model-level effect.

\subsection{More information is not monotonically better}
\label{sec:mechanics-results}

Supplying ground-truth joint type and motion axis on top of the target region and identity lowers action accuracy for every model on GAPartNet, from 0.84 to 0.68, 0.68 to 0.53 and 0.89 to 0.79, and for two of three on the curated UMD stratum, from 1.00 to 0.93 twice. Across six model-stratum pairs, the mechanics-enriched condition never improves on its direct predecessor, \texttt{localization\_shown}. Mechanics were available for every part scored under this condition, so this is not a coverage gap being reported as a decline.

We do not have a mechanism for this and will not invent one from three points per stratum. What it does establish is that this design's information ordering, which is a claim about what a prompt supplies, does not induce the matching ordering in accuracy, so a benchmark of this shape cannot assume more grounding is monotonically better and read a drop as noise. The effect is consistent in direction across every model and both strata, which is the part that makes it worth reporting rather than the size of any single drop.

\subsection{Counterfactual compliance, and where it breaks}
\label{sec:counterfactual-results}

On the frozen 74-pair pool, all 74 pairs are informative in the sense of Section~\ref{sec:action-contrast}, spanning 32 distinct objects. Compliance with the swapped-in part is 0.66, 0.84 and 0.86 for GPT-4o mini, GPT-5 mini and GPT-5, with object-macro rates of 0.68, 0.79 and 0.83 and object-clustered 95\% intervals of $[0.54, 0.79]$, $[0.72, 0.94]$ and $[0.75, 0.96]$. Shortcut rates, where a model answers with the original target's action instead of the one it was asked about, are 0.08, 0.00 and 0.07. Read alone, these show that outputs usually follow the changed named-part request rather than defaulting to the original target's action. They do not establish visual attention to the named part, because the supplied target-part category may itself imply the requested action and no counterfactual no-vision control was run.

The transition table qualifies this sharply. Compliance is 0.66 or better on every observed transition for every model except one: \texttt{push} to \texttt{lift-vertical}, where the three models score 0.33, 0.22 and 0.00 across 9 pairs. GPT-5 fails all nine.

Both available readings are partly right and we report both. GPT-5 emitted \texttt{lift-vertical} twice in 356 predictions across the entire run, and never once on this pool, so the failure is confounded with a label these models are simply reluctant to produce, and a compliance rate cannot cleanly be read as a reasoning measurement on a transition whose target label is nearly absent from the model's output distribution. At the same time, five of GPT-5's nine failures returned specifically the original part's action, which is the shortcut signature rather than an arbitrary miss. The pooled 0.86 would have hidden both facts, which is the argument for the transition table rather than a footnote about it.

\subsection{An open-weight pilot, and a metric with no measurable value}
\label{sec:qwen-pilot}

Under \texttt{localization\_shown}, we ran the open-weight Qwen3-VL-8B against 50 frame-level examples from the unaudited bulk UMD stratum, spanning 43 source tool instances. This is not the curated 28-object OpenAI split: the two runs share no exact image paths and overlap on 11 tool instances, so their scores are not a direct model comparison. Qwen's action accuracy was 0.98 against a majority baseline of 0.32. Mean localization distance was 0.0 at an IoU of 1.0 against a constant-baseline distance of 0.085, and both figures are exact rather than rounded, because the model returned the supplied point verbatim on all 50 examples. Scoring how precisely a model locates a region the prompt already handed it measures whether it can copy a number, which is the same caution Section~\ref{sec:location-not-name} draws from the frontier models under this condition and the reason we do not read a localization figure from it.

Part-selection accuracy here is not measurable in either direction, and three successive versions of this evaluation reported it as though it were. The first gave 1.00, an artifact of the harness supplying the target's identity whenever the condition already contained it and the model's own output omitted it; the raw responses show the model never populated that field on any of the 50. Requiring a genuine selection replaced that with 0.00. The current evaluator, which credits an exact bounding box as real geometric evidence of a selection, and which Section~\ref{sec:location-not-name}'s results depend on doing, replays the same stored predictions to 1.00 again. All three are wrong in the same way: under this condition the only geometric evidence available is the box the prompt supplied, so crediting it scores an echo and withholding credit scores a field the model had no reason to fill. We report this condition's action accuracy and no part-selection figure for it.

A separate audit of this run's coordinate convention, released with the benchmark, found the stored localization distances for \texttt{image\_only}, \texttt{object\_named} and \texttt{part\_named}, all near 0.54, to be a mismatch between the model's documented output convention and the harness's assumption rather than a measurement of anything; they replay to near 0.07. That error and the two the companion evaluation disclosed were each caught by checking a number against something trivial, never by rereading the code that produced it, which is the practice this benchmark is built to make routine rather than occasional.

\section{Limitations}
\label{sec:limitations}

Two of the safeguards above are themselves incomplete in ways worth stating rather than leaving implicit. The curated GAPartNet stratum's target-part bounding boxes were, at an earlier stage of this benchmark, an approximate box centered on a previously verified grasp point rather than an exact reprojection of the part's real three-dimensional bounding box, because the camera pose used to render these objects was computed at render time and not persisted. Recovering that pose and projecting the real box was a straightforward fix and revealed a second, more interesting problem: for most \texttt{slider\_button} targets, and one \texttt{slider\_drawer}, GAPartNet's own per-part three-dimensional bounding box is not expressed in the same coordinate frame as the rendered mesh, so the projected box lands nowhere near the part. A per-link joint-origin correction was tested as the obvious explanation and ruled out, since some of the affected links have a zero joint origin. \texttt{hinge\_door} targets are mostly unaffected. Of the 19 curated objects, 7 (six doors and one drawer) now use a verified exact projection; the other 12 keep the approximate box, labelled as such per object rather than asserted to be solved across the stratum. Whichever measurement error a benchmark catches first depends on which safeguard it builds first, and this one was caught by literally drawing the projected box on the image and looking, the same low-technology check that caught both of the companion evaluation's own errors.

Five caveats bound Section~\ref{sec:results} specifically. First, GPT-4o mini violated the requested bare-JSON format on 113 of 114 GAPartNet predictions, 166 of 168 UMD predictions and 73 of 74 counterfactual predictions, against zero violations for both GPT-5 models. These image-shown violations were fenced JSON responses rather than unusable answers, and the lenient parser recovered a complete semantic response every time; nevertheless, that model's numbers use a recovery path the other two never touched, so its lower counterfactual compliance should not be read as a capability gap until format is controlled. In the separate text-only control, 66 of its 114 outputs violated the strict format; 65 were recoverable fenced objects and one contained an out-of-vocabulary action that was normalized to \texttt{none} and scored incorrect. Second, the curated UMD stratum has \texttt{grasp} as the target action for all 28 objects, so its majority-action baseline is 1.00 and its \texttt{part\_named} scores of 1.00 are exactly that baseline; UMD constrains nothing about action discrimination here and we draw no conclusion from it, reporting it only for the localization and mechanics comparisons. Third, every prompt supplies the same eleven-action vocabulary across both strata, so under the three under-informed conditions models on GAPartNet objects sometimes answer with tool affordances such as \texttt{grasp} or \texttt{wrap-grasp} that the stratum never uses, which depresses \texttt{image\_only} in particular; \texttt{localization\_only}, where GPT-5 kept 15 of 19 answers inside the push and pull family and still only matched the baseline, is much less exposed to this. Fourth, each API condition was sampled once using provider-default decoding. The bootstrap intervals quantify variation across objects, not stochastic variation across repeated model responses, and overlapping intervals are not equivalence tests. Finally, 19 objects, 28 objects and 74 pairs are small, the intervals in Section~\ref{sec:counterfactual-results} are correspondingly wide, and the non-monotonicity in Section~\ref{sec:mechanics-results} rests on the effect never reversing across six model-stratum pairs, five drops and one tie, rather than on any single drop being individually significant.

The bulk, automatically converted strata (600 GAPartNet objects, roughly 105 UMD tool instances at one record per frame) are unaudited, unlike the two curated strata drawn from the companion evaluation's spot-checked objects, and are reported as a separate count for exactly that reason. Multi-viewpoint rendering, open-configuration renders, mask-based localization, and the two more expensive counterfactual constructions, cross-object part compositing and mesh-level part substitution, are deferred rather than built into this version. None of the above changes what is reported in Sections~\ref{sec:design} and~\ref{sec:results}; they bound what those sections can be read to establish, in the same spirit as the companion evaluation's own caution about what naming a part in a prompt does and does not demonstrate.

\section{Reproducibility}
\label{sec:reproducibility}

The benchmark, its converters, its evaluator, and the exact commands used to build every count reported in Section~\ref{sec:design} are released at the repository above. Every dataset conversion is deterministic and auditable against its source annotation rather than inferred from geometry, and the evaluator's manifest tool reports per-file statistics, including a check that every referenced image actually exists, for any combination of the files described here. Every rendered image the action-contrast pool depends on is additionally recorded in a checksummed manifest, object id, sha256, the exact render call, and the environment's package versions, so its construction stays auditable independent of which machine produced the images.

\section{Conclusion}

A single accuracy number, or a single before-and-after comparison against it, cannot distinguish a model that cannot find the right part from one that finds it but maps it to the wrong action, or from one that never needed the image because the supplied category already gave the answer away. GroundBench is built to keep those explanations apart: a branch-and-merge condition design that changes one controlled information bundle per edge, a counterfactual re-ask that requests a real noncanonical part, and an evaluator whose thresholds are checked against the same trivial baselines that caught two real errors in the evaluation it extends. Across three OpenAI models, the benchmark-supplied target region without the part name does not beat a constant answer, while the name condition does. Every above-baseline gain occurs where the supplied target-part category itself implies the answer in this curated set, and GPT-5's no-vision result is consistent with category-to-action association contributing substantially to the companion evaluation's headline effect. GPT-4o mini disagrees on one condition, so the conclusion is evidence for that explanation rather than a model-universal causal estimate. Supplying real mechanics on top of the target region and identity also failed to behave as expected: across both strata, it produced five drops and one tie rather than an improvement. These findings make the benchmark most useful as a diagnostic instrument---one that exposes which supplied information changes behavior and which apparently successful metrics can be satisfied without the capability they seem to name.

% Bibliography for "Part Grounding, Not Action Knowledge".
%
% Drop-in replacement for the thebibliography block in paper.tex. Either paste
% this in place of that block, or keep it separate and pull it in with
% \input{references} where the block currently sits.
%
% Author lists and venues were taken from the arXiv API rather than copied from
% search results, so the entries below are the published ones. Two are worth
% noting: ManipBench is CoRL 2025 (not 2024, which some secondary sources say),
% and Jones et al. is AAMAS 2026 main track rather than a preprint only.

\end{document}